%% file: 0.main.tex
\documentclass{article}

\usepackage{microtype}
\usepackage{graphicx}
\usepackage{subcaption}
\usepackage{booktabs}
\usepackage[utf8]{inputenc} 
\usepackage[T1]{fontenc}    
\usepackage{hyperref}       
\usepackage{url}            
\usepackage{amsfonts}       
\usepackage{nicefrac}       
\usepackage{xcolor}         
\usepackage{amssymb}
\usepackage{multirow}
\usepackage[normalem]{ulem}
\useunder{\uline}{\ul}{}
\usepackage{makecell}
\usepackage{enumitem}
\usepackage{amsmath}
\usepackage{float}
\usepackage{placeins}
\usepackage[most]{tcolorbox}
\usepackage{minted}
\usepackage{hyperref}

\usepackage[preprint]{icml2026}

\usepackage{amsmath}
\usepackage{amssymb}
\usepackage{mathtools}
\usepackage{amsthm}
\usepackage{multirow}
\usepackage{makecell}

\usepackage[capitalize,noabbrev]{cleveref}

\theoremstyle{plain}

\theoremstyle{definition}

\theoremstyle{remark}

\usepackage[table]{xcolor}

\newcommand{\tagGetRef}[1]{\textcolor{teal}{\textbf{<#1>}}}
\newcommand{\tagSearch}[1]{\textcolor{orange!80!black}{\textbf{<#1>}}}
\newcommand{\tagAnalyse}[1]{\textcolor{violet!70!black}{\textbf{<#1>}}}
\newcommand{\tagIdeation}[1]{\textcolor{blue!60!black}{\textbf{<#1>}}}
\newcommand{\tagResult}[1]{\textcolor{brown!60!black}{\textbf{<#1>}}}

\newcommand{\resultblock}[1]{%
  \colorbox{yellow!20}{%
    \parbox{\dimexpr\linewidth-2\fboxsep}{#1}%
  }%
}

\usepackage[textsize=tiny]{todonotes}

\begin{document}

\twocolumn[
  \icmltitle{Agentic-Ideation: Sample Efficient Agentic Trajectories Synthesis for Scientific Ideation Agents}




  \icmlsetsymbol{equal}{*}

  \begin{icmlauthorlist}
    \icmlauthor{Keyu Zhao}{tsinghua}
    \icmlauthor{Lingyan Kong}{tsinghua}
    \icmlauthor{Fengli Xu}{tsinghua}
    \icmlauthor{Yong Li}{tsinghua}
  \end{icmlauthorlist}

  \icmlaffiliation{tsinghua}{Department of Electronic Engineering, Tsinghua University}

  \icmlcorrespondingauthor{Fengli Xu}{fenglixu@tsinghua.edu.cn}

  \icmlkeywords{Machine Learning, ICML}

  \vskip 0.3in
]



\printAffiliationsAndNotice{}  

\begin{abstract}
Ideation plays a pivotal role in scientific discovery. Recent LLM, especially AI Scientist systems, show promising potential for automated ideation.
However, existing approaches predominantly rely on pre-defined agentic workflows. This constraint severely limits the flexibility required to navigate the vast search space of scientific literature and the complex action space of research reasoning. 
Recently, training Agentic LLMs has emerged as a promising direction, offering flexible reasoning frameworks and the capability for autonomous tool utilization. However, there remains a non-trivial challenge: applying previous agentic data synthesis methods to scientific ideation suffers from prohibitively high data synthesis cost.
To bridge this gap, we propose Agentic-Ideation, a novel framework comprising an automated trajectory synthesis pipeline and a specialized agentic LLM trained for scientific ideation.
Specifically, we first define a comprehensive tool space incorporating three external tools and three cognitive tools. 
Then we introduce an Oracle-Guided Data Synthesis strategy. By leveraging a reference idea as oracle guidance, this approach steers the multi-agent system to efficiently reconstruct the logical reasoning and tool invocation paths, transforming aimless trial-and-error into directed trajectory generation.
Finally, we train the agent on these synthesized trajectories, employing a masking strategy on tool execution results. This ensures the model focuses on decision-making logic without interference from external feedback.
Experimental results demonstrate that our method outperforms the SOTA workflow-based baseline by \textbf{11.91\%} in overall quality. Furthermore, our approach improves the sample efficiency of high-quality data synthesis by \textbf{over 10$\times$}.
\end{abstract}


\input{1.intro}

\input{2.Related}
\input{3.Methods}
\input{4.Experiments}
\input{5.Conclusion}

\section*{Impact Statements}
This work significantly advances the field of automated scientific discovery by addressing the critical bottleneck of data scarcity. By enabling the efficient synthesis of high-quality agentic trajectories, our framework paves the way for developing autonomous AI scientists capable of grounded reasoning and reliable innovation. This advancement holds the potential to accelerate the pace of research across diverse scientific disciplines, democratizing access to high-level ideation capabilities and fostering human-AI collaboration in solving complex scientific challenges.

\bibliography{reference}
\bibliographystyle{icml2026}

\newpage
\appendix

\input{6.appendix}
\end{document}

%% file: 1.intro.tex
\section{Introduction}
Scientific ideation serves as the genesis of the entire research lifecycle, fundamentally determining the novelty and potential impact of downstream discoveries. Recently, the rapid evolution of Large Language Models (LLMs) has catalyzed the vision of an "AI Scientist"~\cite{tang2025ai, lu2024ai, schmidgall2025agent}, capable of automating complex research tasks ranging from initial idea generation~\cite{yu2024researchtown, pu2025piflow} to final manuscript preparation~\cite{wang2024autosurvey,yan2025surveyforge, liang2025surveyx}.

Despite this priority, previous approaches predominantly rely on pre-defined workflows~\cite{li2024chain, zhou2025trustresearcher}. These methods typically constrain Large Language Models (LLMs) to execute pre-defined, linear sequences, such as literature retrieval followed by template filling. Crucially, the efficacy of these rigid pipelines depends heavily on complex, human-designed prompts meticulously crafted to guide the model through each step. This reliance on static, hand-crafted engineering restricts flexibility~\cite{Shang2024AgentSquareAL, Hu2024AutomatedDO}, failing to navigate the vast search space of literature where researchers must adaptively select tools and refine reasoning. To transcend these limitations, training Agentic LLMs has emerged as a promising frontier. Unlike static workflows, agentic models offer a flexible reasoning framework with the potential to autonomously orchestrate tools and adaptively navigate the research space without being bound by brittle scripts~\cite{Jin2025SearchR1TL, Chen2025ReSearchLT}.

However, training such agentic models is non-trivial, primarily due to the challenge of efficiently synthesizing high-quality agentic trajectory data. Existing data synthesis frameworks for agents generally focus on tasks with definite answers, such as multi-hop Question Answering (QA)~\cite{Tao2025WebShaperAD, Li2025WebSailorV2BT}. In these scenarios, high-quality trajectories are typically filtered via rejection sampling~\cite{Li2025WebSailorNS}, where the correctness of the final answer serves as a clear ground truth signal to validate the reasoning path. Unfortunately, this paradigm is ill-suited for scientific ideation. Unlike QA, ideation involves a vast and open-ended answer space where no single "correct" solution exists~\cite{Zhang2025ScientificJD, Calic2020SubjectiveSS}. Consequently, applying standard rejection sampling leads to prohibitively low sample efficiency, as unguided models struggle to navigate this immense search space to yield high-quality ideas, rendering the data synthesis process computationally expensive and ineffective.

To tackle these challenges, we introduce Agentic-Ideation, a comprehensive framework designed for efficient agentic data synthesis and robust model training. We first meticulously define the agent's hybrid tool space, which comprises three external tools and three thinking tools. The external tools, including \textit{Search}, \textit{Get\_References} and \textit{Get\_Cited}, empower the agent to access external APIs, thereby proactively expanding its knowledge boundaries beyond its internal parameters. Complementing these, the thinking tools, including \textit{Analyse\_Gap}, \textit{Ideation} and \textit{Reflection}, enable the agent to perform critical cognitive tasks, such as identifying research voids and iteratively refining ideas. To synthesize high-quality trajectories based on this action space, we construct a hierarchical multi-agent system consisting of a Planner, a Controller, and six specialized tool agents. In this architecture, the Planner orchestrates the high-level ideation strategy, while the Controller manages fine-grained tool selection and execution at each step.

Notably, we endow agents with an oracle view of the high-quality "reference idea," acting as a navigational beacon to maximize sample efficiency. Instead of aimlessly exploring the vast action space, this guidance transforms the synthesis process from a stochastic search into a directed trajectory reconstruction. By leveraging this "foresight," the system can generate a valid, logically coherent reasoning path in a single pass. This fundamentally circumvents the massive computational waste inherent in traditional rejection sampling, where valid trajectories are sparsely distilled only after exhaustive, undirected "hindsight" trial-and-error. Finally, we utilize these high-quality trajectories to train the target agent, equipping it with the capability to perform grounded reasoning and autonomous exploration for open-ended scientific ideation. During this phase, we apply a strategic masking technique to the results returned by external tools. This critical design forces the model to focus on the reasoning behind tool invocation rather than memorizing specific return values, thereby avoiding interference from external feedback during the learning process and ensuring the generalization and robustness of the trained agent.

Extensive experiments demonstrate the superiority of our proposed framework. Quantitatively, our trained agentic LLM consistently outperforms all competitive baselines, achieving an 11.91\% improvement in overall score compared to the state-of-the-art method, which empirically validates the effectiveness of our proposed agentic training paradigm. In terms of data construction, our approach achieves a remarkable efficiency boost, accelerating the synthesis of high-quality agentic trajectories by over tenfold compared to previous rejection sampling strategies. Furthermore, a comprehensive human evaluation confirms that the ideas generated by our agent exhibit higher novelty, feasibility, and scientific value, validating the practical utility of our method in real-world research scenarios.

 Our contributions can be summarized as follows:
\vspace{-2mm}

\begin{enumerate}
\setlength{\itemsep}{1pt} 
\setlength{\parskip}{0pt} 
\setlength{\parsep}{0pt} 
    \item We pioneer the application of agentic LLMs to scientific ideation, offering a flexible and autonomous alternative to existing rigid, workflow-based approaches. This paradigm shift enables the model to adaptively navigate the complex search space of scientific ideas.
    
    \item We propose a novel oracle-guided multi-agent data synthesis framework. By leveraging a global view to guide trajectory generation, we address the challenge of data scarcity and significantly enhance sample efficiency in the open-ended ideation domain.
    
    \item We achieve SOTA results across all metrics, outperforming strong baselines and improving synthesis efficiency by over tenfold. Furthermore, human evaluation confirms our method generates more novel and feasible ideas.
\end{enumerate}

%% file: 2.Related.tex
\section{Related Works}

\subsection{Scientific Ideation}

\begin{figure*}[t]
    \centering
    \includegraphics[width=\linewidth]{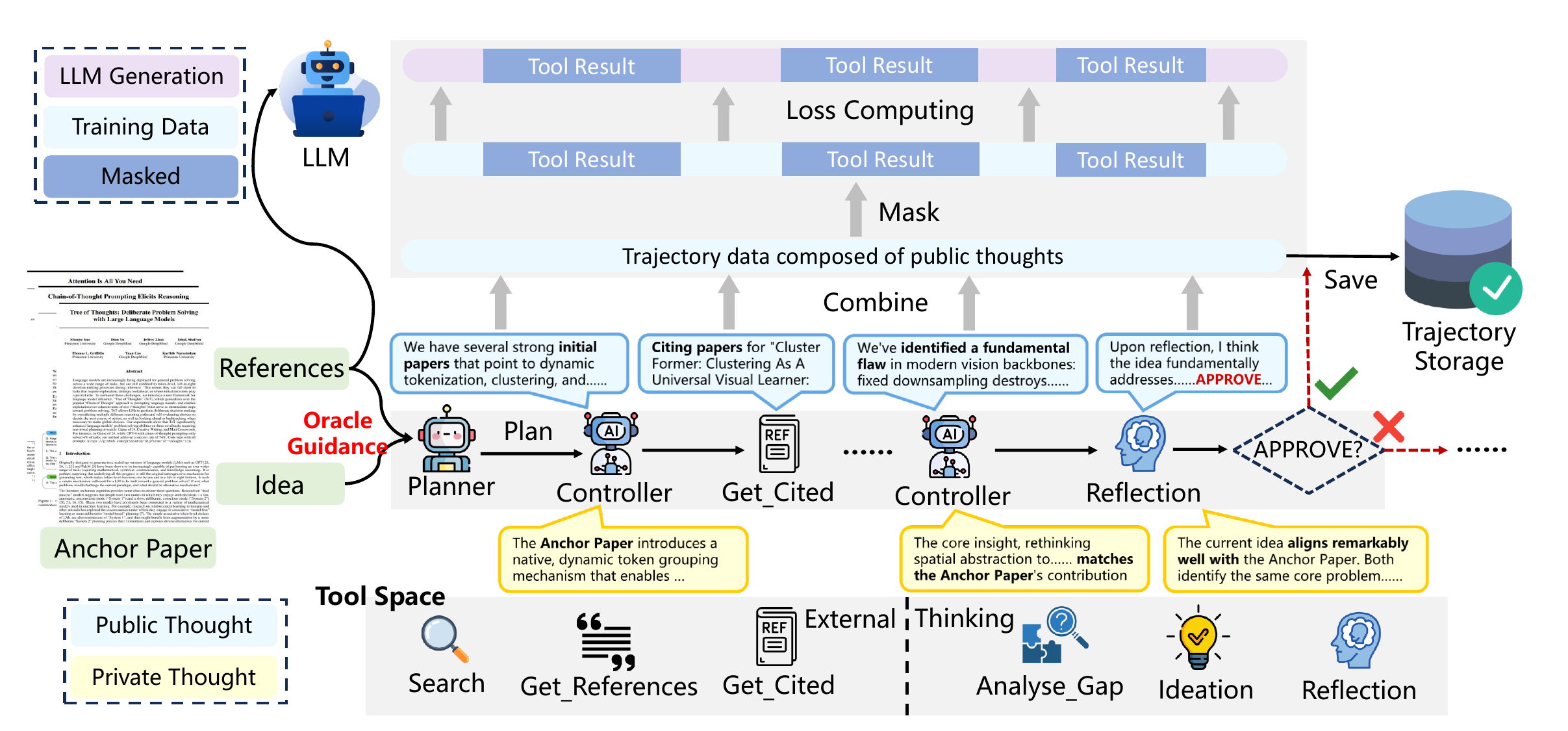}
    \caption{Overview of the proposed AgenticIdeation includes \textbf{Oracle-Guided Agentic Data Synthesis} and \textbf{Agentic Supervised Fine-Tuning}. The system utilizes Oracle guidance to synthesize high-quality research trajectories, which are subsequently used to fine-tune the model with a masking strategy applied to tool execution results.}
    \label{fig1}
    \vspace{-2mm}
\end{figure*}

The pursuit of AI Scientists envisions systems capable of automating the entire research lifecycle~\cite{yamada2025ai, gottweis2025towards, schmidgall2025agentrxiv}. Central to this vision is Scientific Ideation, the foundational upstream phase that determines the novelty and validity of downstream experimentation~\cite{Novikov2025AlphaEvolveAC, RomeraParedes2023MathematicalDF}. While recent works have explored various generation strategies. ResearchAgent~\cite{Baek2024ResearchAgentIR} augments idea generation by utilizing a scientific knowledge graph and an iterative peer review mechanism, while SciPiP~\cite{Wang2024SciPIPAL} employs a dual-path framework combining retrieval-based and generative approaches to balance novelty and feasibility. Focusing on uniqueness, SciMON~\cite{Wang2023SciMONSI} explicitly optimizes for novelty by iteratively retrieving inspirations from past literature to refine hypotheses, whereas VirSci~\cite{Su2024ManyHA} simulates a virtual research team where multiple agents collaboratively propose and evaluate ideas. Despite their specific innovations, these approaches share a critical limitation: they rely on fixed, static workflows driven by complex human-designed prompts. This rigidity constrains the model to follow pre-defined procedures, preventing the autonomous flexibility required for truly open-ended scientific exploration.

\subsection{Agentic Data Synthesis}
Distinct from rigid workflow-based systems, Agentic LLMs are defined by their autonomy in planning and invoking external tools to solve open-ended problems. To equip models with these capabilities, Agentic Data Synthesis shifts the training paradigm from learning static text to mimicking dynamic interaction trajectories, explicitly capturing the interplay between reasoning, tool execution, and environmental feedback. Recent works employ diverse strategies to synthesize agentic training data. AgentScaler~\cite{Fang2025TowardsGA} automates experience accumulation by formalizing environments as read-write databases to generate interaction logs, while WebDancer~\cite{Wu2025WebDancerTA} captures end-to-end browser trajectories specifically tailored for complex "Deep Research" retrieval. Addressing reasoning depth, WebSailor~\cite{Li2025WebSailorNS} synthesizes high-difficulty tasks through graph sampling and information obfuscation, whereas WebWatcher~\cite{Geng2025WebWatcherBN} extends synthesis to the multimodal domain by integrating visual web contexts. However, these methods typically depend on outcome-based filtering mechanisms (e.g., rejection sampling or task completion verification) to ensure data quality. In the vast, open-ended answer space of scientific ideation, where unique ground truth or immediate success signals are absent, such strategies suffer from extremely low sample efficiency, as unguided exploration fails to consistently yield high-quality outcomes without explicit directional guidance.

%% file: 3.Methods.tex
\section{Methodology}

The Agentic-Ideation framework streamlines autonomous scientific discovery through three integrated stages. First, we establish a hybrid tool space that combines external retrieval utilities with internal cognitive tools to support complex reasoning. Second, to overcome data scarcity, we employ an oracle-guided multi-agent system where a Planner and Controller, guided by "reference ideas," efficiently synthesize high-fidelity trajectories without relying on inefficient trial-and-error. Finally, the target LLM is trained via Supervised Fine-Tuning, where tool outputs are masked to ensure the model learns robust decision-making logic rather than memorizing external feedback. The overall framework of Agentic-Ideation are presented in Figure~\ref{fig1}.

\subsection{Agentic Tool Space Definition}
Scientific ideation is inherently an iterative cognitive process that oscillates between the extensive accumulation of domain knowledge and the intensive synthesis of novel ideas. To faithfully model this complex behavior, we define a comprehensive Agentic Tool Space that bridges the gap between retrieving external evidence and performing internal reasoning. We formalize this space as $\mathcal{T} = \mathcal{T}_{ext} \cup \mathcal{T}_{think}$, where $\mathcal{T}_{ext}$ denotes external retrieval tools and $\mathcal{T}_{think}$ represents internal reasoning tools. Let $\mathcal{H}$ denote the current interaction history.

\textbf{External Tools ($\mathcal{T}_{ext}$)} These tools interface with the academic database. Notably, to simulate realistic constraints, only the search tool returns full details, while citation tools provide broad, shallow context.

\begin{itemize}
\setlength{\itemsep}{1pt} 
\setlength{\parskip}{0pt} 
\setlength{\parsep}{0pt} 
    \item $\textit{Search}(q) \rightarrow \mathcal{D}_{meta}$: Queries the database with keywords $q$. It returns a set of papers $\mathcal{D}_{meta}$ containing comprehensive metadata (e.g., abstract, citation count, publication year), serving as the primary source for detailed information acquisition.
    
    \item $\textit{Get\_References}(d_{id}) \rightarrow \mathcal{L}_{titles}$: Retrieves the bibliography of a target paper $d_{id}$. It returns a list of titles only, allowing the agent to quickly scan the foundational roots of a work without overloading the context.
    
    \item $\textit{Get\_Cited}(d_{id}) \rightarrow \mathcal{L}_{titles}$: Identifies papers that have cited $d_{id}$. Similarly, it returns a list of titles only, enabling the agent to efficiently map the downstream evolution and future trends related to the target paper.
\end{itemize}

\textbf{Thinking Tools ($\mathcal{T}_{think}$)} These tools model latent cognitive steps, mapping the context $\mathcal{H}$ to explicit reasoning outputs $o$.

\begin{itemize}
\setlength{\itemsep}{1pt} 
\setlength{\parskip}{0pt} 
\setlength{\parsep}{0pt} 
    \item $\textit{Analyse\_Gap}(\mathcal{H}) \rightarrow o_{gap}$: Synthesizes the retrieved information in $\mathcal{H}$ to identify logical inconsistencies or underexplored intersections, explicitly outputting a "research void" $o_{gap}$.
    
    \item $\textit{Ideation}(\mathcal{H}, o_{gap}) \rightarrow o_{idea}$: The core generative action. Based on the identified gap $o_{gap}$, it proposes a novel scientific idea $o_{idea}$, detailing the innovation and expected advantages.
    
    \item $\textit{Reflection}(\mathcal{H}, o_{idea}) \rightarrow o_{eval}$: A self-correction mechanism that critically evaluates the quality of generated idea $o_{idea}$, outputting a critique $o_{eval}$ to trigger refinement if necessary.
\end{itemize}

\subsection{Oracle-Guided Agentic Data Synthesis}

To systematically synthesize agentic trajectories, we design a hierarchical multi-agent system operating on the defined tool space $\mathcal{T}$. The process initiates by selecting a scientific paper, termed the "Anchor Paper," and extracting its bibliography to serve as the system input. The system then decomposes the ideation process into strategic planning and tactical execution, formalized as follows:

\begin{itemize}
\setlength{\itemsep}{1pt} 
\setlength{\parskip}{0pt} 
\setlength{\parsep}{0pt} 
    \item \textbf{Planner} ($A_{plan}$): Acts as the architect. Given the initial context $\mathcal{H}_0$ (derived from the anchor paper's references), the planner generates a high-level research roadmap $P$ to structure the exploration:
    $$P = A_{plan}(\mathcal{H}_0)$$
    
    \item \textbf{Controller} ($A_{ctrl}$): Acts as the executor. At each time step $t$, guided by the roadmap $P$, the controller processes the current interaction history $\mathcal{H}_t$ to generate a reasoning thought $Th_t$ and a subsequent action $a_t$:
    $$(Th_t, a_t) = A_{ctrl}(\mathcal{H}_t, P, I_{ref})$$
    where $a_t \in \mathcal{T}$ denotes the specific tool invoked to acquire new information or perform reasoning.
\end{itemize}

\begin{itemize}
\setlength{\itemsep}{1pt} 
\setlength{\parskip}{0pt} 
\setlength{\parsep}{0pt} 
    \item \textbf{Private Thought} ($th_{priv}^{(t)}$): Represents the \textbf{Oracle View}. It explicitly utilizes $I_{ref}$ to perform a gap analysis between the current state and the target goal. This stream calculates the optimal tool $a_t$ required to bridge the gap, effectively collapsing the search space by providing directional foresight.
    
    \item \textbf{Public Thought} ($th_{pub}^{(t)}$): Represents the \textbf{Researcher View}. It rationalizes the decision $a_t$ using only the currently available history $\mathcal{H}_t$, intentionally masking the existence of $I_{ref}$. This ensures the resulting reasoning trace mimics a natural, autonomous exploration process suitable for training.
    
\end{itemize}

\begin{figure}[H]
\vspace{-3mm}
    \centering
    \includegraphics[width=\linewidth]{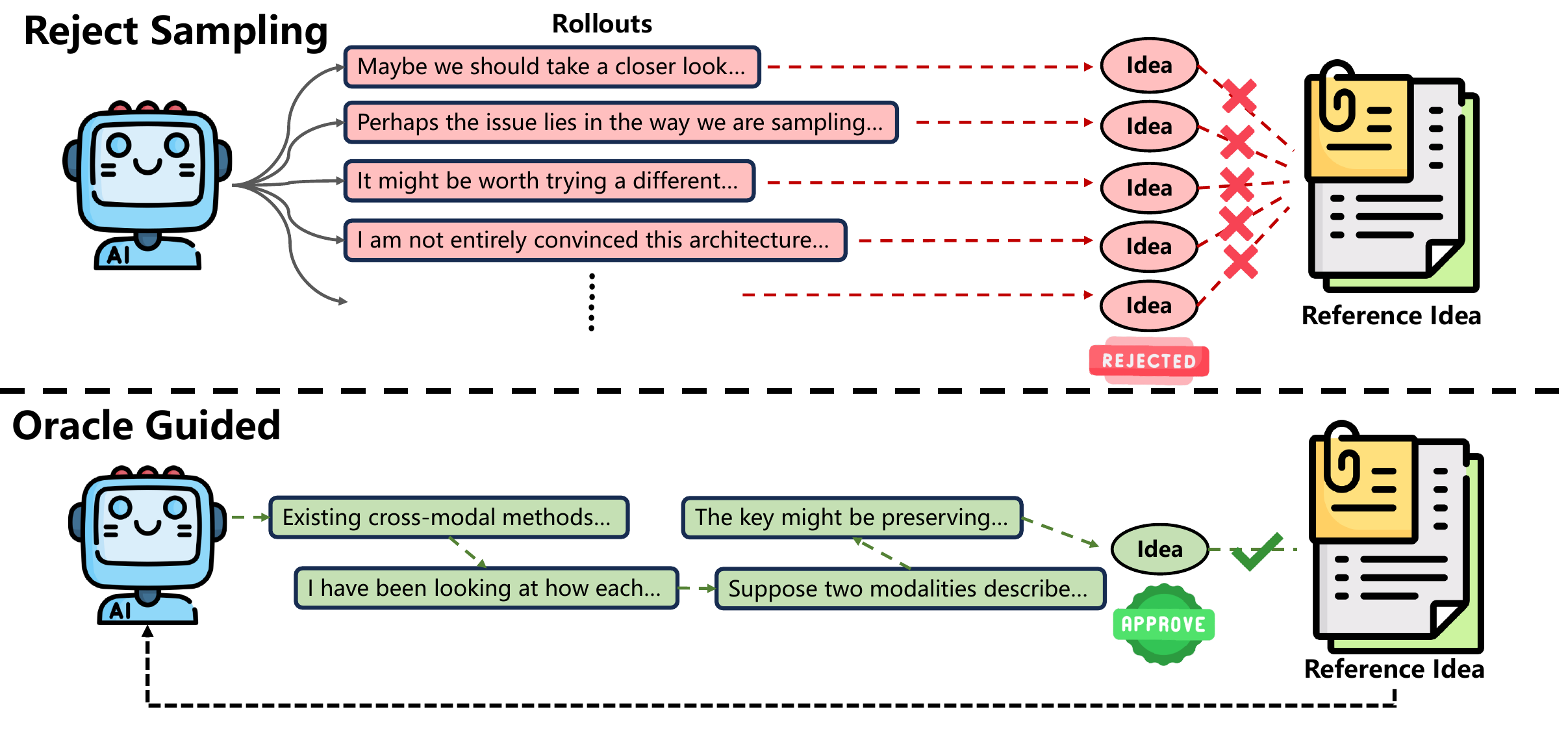}
    \caption{Comparsion between \textbf{Reject Sampling} and our \textbf{Oracle Guided strategy}.}
    \label{fig2}
    \vspace{-2mm}
\end{figure}

However, utilizing this system directly for data synthesis suffers from severe inefficiency, as shown in Figure~\ref{fig2}. A naive approach relies on rejection sampling as a quality filter, where a generated trajectory $\tau$ is retained only if its final idea $o_{idea}$ matches the anchor paper's core idea. In the vast, open-ended search space of scientific ideation, an unguided agent rarely converges spontaneously on a specific target, resulting in a prohibitively high rejection rate. To address this, we propose an Oracle-Guided Strategy by injecting the anchor paper's core idea, denoted as the Reference Idea ($I_{ref}$), into the inference process. Specifically, we implement a Dual-Thought Mechanism where the controller simultaneously generates two distinct cognitive streams. Formally, the policy at step $t$ is defined as:
$$(th_{priv}^{(t)}, th_{pub}^{(t)}, a_t) = A_{ctrl}(\mathcal{H}_t, P, I_{ref})$$
The two thought components function as follows:

The synthesis process concludes when the agent invokes the $\textit{Reflection}$ tool. The system evaluates the semantic alignment between the generated idea $o_{idea}$ and $I_{ref}$. Only trajectories where the generated idea successfully recovers the reference idea are preserved for the subsequent training phase.

\subsection{Agentic Supervised Fine-Tuning}
To transform the diverse interaction traces generated by our multi-agent system into a format suitable for LLM training, we employ a structured linearization strategy. The input to the model consists of a system prompt defining the tool space $\mathcal{T}$ and the user query containing the references of the anchor paper. Crucially, to ensure the trained agent learns autonomous exploration rather than "cheating," we exclusively retain the Public Thought ($th_{pub}$) from the controller's dual-stream output as the primary reasoning chain, discarding the Private Thought ($th_{priv}$) and the Oracle's guidance. The training sequence is rigorously organized using special tokens: external tool queries are encapsulated as \textit{<Tool\_Name>}Query\textit{</Tool\_Name>}, environmental observations are wrapped in \textit{<Result>}Output\textit{</Result>}, and internal cognitive actions are formatted as \textit{<Tool\_Name>}Content\textit{</Tool\_Name>}, culminating in the final output marked by \textit{<Answer>}Final Idea\textit{</Answer>}. The structure of the trajectory is shown in the Figure~\ref{trajectory}.

\begin{figure}[H]
\vspace{-4mm}
\centering
\begin{tcolorbox}[
    breakable,
    title={Trajectory},
    colback=white,
    boxrule=0.5pt,
    left=4pt, right=4pt, top=3pt, bottom=3pt,
    fonttitle=\bfseries\sffamily,
    width=\linewidth 
]
I should first gather relevant literature...

\tagGetRef{Get\_References}

(Tool Call Query)

\tagGetRef{/Get\_References}

\resultblock{
\tagResult{Result}

(Tool Call Result)

\tagResult{/Result}
}

I need to dig deeper into how it’s used in practice...

\tagSearch{Search}

(Tool Call Query)

\tagSearch{/Search}

\resultblock{
\tagResult{Result}

(Tool Call Result)

\tagResult{/Result}
}

Let me reflect on what’s still missing or underexplored.

\tagAnalyse{Analyse\_Gap}

(Analysing current gap)

\tagAnalyse{/Analyse\_Gap}

I now have enough context to propose a new idea.

\tagIdeation{Ideation}

(Generating Idea...)

\tagIdeation{/Ideation}
\end{tcolorbox}

\caption{The structure of an agentic trajectory training data.}
\label{trajectory}
\vspace{-6pt}
\end{figure}

For model training, we utilize a Supervised Fine-Tuning (SFT) approach. Following ~\cite{Chen2023FireActTL, Jin2025SearchR1TL, Zhang2025AgentMI}, we mask tool execution results during loss calculation to ensure training performance and robustness. This design prevents the model from memorizing deterministic environmental feedback, forcing it to focus solely on reasoning and decision-making. The optimization objective is defined as:

$$\mathcal{L}(\theta) = - \frac{1}{\sum_{t=1}^{|X|} M_t} \sum_{t=1}^{|X|} M_t \cdot \log P_\theta (x_t \mid x_{<t})$$

where $X$ represents the complete linearized token sequence, and $x_t$ denotes the $t$-th token. $x_{<t}$ represents the context preceding $x_t$, and $P_\theta$ is the conditional probability distribution parameterized by the model weights $\theta$. The binary mask $M_t \in \{0, 1\}$ serves as the indicator function: $M_t=1$ if $x_t$ belongs to the agent's generated actions (thoughts or tool calls), and $M_t=0$ if $x_t$ belongs to system instructions or environmental feedback (tokens within <Result>...</Result>). This ensures gradients are back-propagated exclusively through the agent's autonomous decisions.

%% file: 4.Experiments.tex
\section{Experiments}

\subsection{Experimental Setup}

\textbf{Baselines.}
To comprehensively evaluate the performance of our proposed framework, we compare Agentic-Ideation against four representative workflow-based ideation systems and the foundation model itself.

\begin{itemize}[left=0pt]
\setlength{\itemsep}{1pt} 
\setlength{\parskip}{0pt} 
\setlength{\parsep}{0pt} 
    \item \textbf{ResearchAgent~\cite{Baek2024ResearchAgentIR}:} An iterative framework that augments ideation using a scientific knowledge store. It refines ideas through simulated peer reviews generated by "ReviewingAgents."
    
    \item \textbf{SciPIP~\cite{Wang2024SciPIPAL}:} A dual-path mechanism that integrates retrieval-based knowledge extraction with generative brainstorming to balance the novelty and feasibility of proposed ideas.

    \item \textbf{SciMON~\cite{Wang2023SciMONSI}:} A system that explicitly optimizes for novelty. It iteratively retrieves "inspirations" from past literature and refines generated ideas until they sufficiently diverge from retrieved prior works.

    \item \textbf{VirSci~\cite{Su2024ManyHA}:} A multi-agent collaboration framework ("Many Heads Are Better Than One") that simulates a team of virtual scientists to collaboratively propose and evaluate ideas through structured discussion.

    \item \textbf{Qwen3-8B~\cite{Yang2025Qwen3TR}:} The backbone model used in our method. We evaluate its zero-shot performance using standard prompting to quantify the specific gains achieved by our agentic fine-tuning.
    
\end{itemize}

\setlength{\tabcolsep}{0.5mm} 
\begin{table*}
    \centering
    \caption{Performance of Agentic-Ideation compared to state-of-the-art baselines across NLP, CV, and Other domains. Bold and underline indicate the best and second-best performance, respectively. $^{*}$ denotes statistical significance ($p\text{-value} < 0.05$) compared to the second-best result.}
    \label{table1}
    \fontsize{12}{10}\selectfont
    \resizebox{\textwidth}{!}{
    \begin{tabular}{ccccc|cccc|cccc|cccc}
    \toprule
    
    \multirow{2}{*}{Method}& \multicolumn{4}{c|}{\textbf{NLP}}& \multicolumn{4}{c|}{\textbf{CV}}& \multicolumn{4}{c|}{\textbf{Others}}& \multicolumn{4}{c}{\textbf{Overall}}\\
    
    & Nov. & Sig. & Fea.& Overall  & Nov. & Sig. & Fea.& Overall  & Nov. & Sig. & Fea.& Overall  & Nov. & Sig. & Fea.& Overall\\ 
    \midrule
    ResearchAgent &5.67& \underline{7.13}& 5.64& \underline{6.01}& 6.02& 6.84& 5.61& 5.94& 6.05& 6.86& 5.54& 5.86&5.91 & 6.94 & 5.60 & 5.94\\ 
    SciPIP &\underline{5.82} & 6.92& \underline{5.71}& 5.97& \underline{6.16}& \underline{7.03}& 5.54& \underline{6.02}& 6.18& 6.94& \underline{5.63}& 5.96& \underline{6.05}& 6.96& \underline{5.63}&5.88\\ 
    SciMON  &5.72 &7.05 &5.54 &5.87 &6.09 &6.93 &\underline{5.62} &5.99 &\underline{6.28} &\underline{7.02} &5.59 &6.03 &6.03 &\underline{7.00} &5.58 &\underline{5.96}\\ 
    VirSci  &5.71 &6.89 &5.67 &5.93 &5.93 &6.91 &5.50 &5.94 &6.22 &6.86 &5.38 &6.01 &5.95 &6.89 &5.52 &5.81\\ 
    Qwen-3-8B &5.62 &7.08 &5.59 &5.90 &6.07 &6.89 &5.46 &5.85 &6.01 &6.91 &5.24 &\underline{6.10} &5.90 &6.96 &5.43 &5.90\\ 
    \midrule
    Agentic-Ideation &\textbf{6.56$^{*}$} &\textbf{7.88$^{*}$} &\textbf{6.21$^{*}$} &\textbf{6.64$^{*}$} &\textbf{6.74$^{*}$} &\textbf{7.65$^{*}$} &\textbf{6.13$^{*}$} &\textbf{6.69$^{*}$} &\textbf{6.92$^{*}$} &\textbf{7.82$^{*}$} &\textbf{6.02$^{*}$} &\textbf{6.69$^{*}$} &\textbf{6.74$^{*}$} &\textbf{7.78$^{*}$} &\textbf{6.12$^{*}$} &\textbf{6.67$^{*}$}\\
    \midrule
    \fontsize{12}{5}\selectfont Improvement\scalebox{1}{$\uparrow$} & 12.71\% &10.52\% &8.76\% &10.48\% &9.42\% &8.82\% &9.07\% &11.13\% &10.19\% &11.40\% &6.93\% &9.67\% &11.40\% &11.14\% &8.70\% &11.91\%\\ 
    \bottomrule
    \end{tabular}
    }
    
\end{table*}

\textbf{Dataset.}
We construct a high-quality dataset derived from accepted papers at three top-tier machine learning conferences: ICLR 2025, ICML 2025, and NeurIPS 2025. These papers serve as "Anchor Papers." To ensure data integrity, we filter out papers where the abstract or bibliography is inaccessible. To facilitate fine-grained evaluation, we employ GPT-5-mini to classify these papers into three distinct domains: Natural Language Processing (NLP), Computer Vision (CV), and Others.

For the test set, we focus on the most recently published conference, NeurIPS 2025, to assess the model's ability to generate cutting-edge ideas. Specifically, we randomly sample 100 papers from each category (NLP, CV, Others), creating a balanced evaluation benchmark of 300 papers. For each anchor paper in both training and testing sets, we randomly select 10 references from its bibliography. These references serve as the sole input context, challenging the models to generate high-quality scientific ideas based strictly on this foundational knowledge.

\textbf{Evaluation Metrics.} 
Following established protocols in automated scientific discovery, we employ four distinct metrics to evaluate the generated ideas. These metrics are scored on a Likert scale (e.g., 1-10). More details can be found in the Appendix ~\ref{metrics}.

\begin{itemize}
\setlength{\itemsep}{1pt} 
\setlength{\parskip}{0pt} 
\setlength{\parsep}{0pt} 
    \item \textbf{Novelty:} Measures the degree of innovation and distinctiveness of the idea compared to existing literature.
    
    \item \textbf{Significance:} Assesses the potential impact and scientific value of the proposed idea to the research community.

    \item \textbf{Feasibility:} Evaluates the technical plausibility and implementation practicality of the proposed method.

    \item \textbf{Overall:} A holistic score reflecting the general quality and completeness of the scientific idea.
    
\end{itemize}

\textbf{Implementation Details.} 
During the data synthesis phase, to ensure the highest quality of "Oracle" guidance, we utilize the powerful Qwen3-235B-A22B-Instruct-2507~\cite{Yang2025Qwen3TR} as the underlying LLM for all agents (Planner, Controller, and Tools) within the oracle-guided multi-agent system. For the Agentic training phase, we employ Qwen3-8B as the backbone for the target agentic LLM, which is fine-tuned on a dataset comprising 4,646 synthesized trajectories. Please refer to Appendix \ref{implementation} for further implementation details.

To ensure a rigorous comparison, all baseline methods (ResearchAgent, SciPIP, SciMON, VirSci) are implemented using GPT-4o as the underlying engine, representing a strong closed-source benchmark.

Given the complexity of scientific ideation, we adopt a "Panel of LLM Judges" approach to mitigate single-model bias. We employ four state-of-the-art models as evaluators: GPT-5.2, Gemini-3-Pro-Preview, Claude-Sonnet-4.5, and Qwen3-Max. The reported scores are the averaged results across these four judges.

\subsection{Main Result}

Table \ref{table1} presents the comprehensive evaluation results. Agentic-Ideation achieves state-of-the-art performance, consistently outperforming all baselines. Specifically, our method achieves an Overall score of 6.67, surpassing the best-performing baseline by 11.91\%. We observe significant gains across Novelty (+11.40\%), Significance (+11.14\%), and Feasibility (+8.70\%), demonstrating the robustness of our approach across NLP, CV, and Other domains.

\textbf{Comparison with Workflow-based Baselines.} Compared to workflow-based systems (e.g., ResearchAgent, SciPIP), Agentic-Ideation exhibits superior adaptability. While baselines are constrained by rigid, pre-defined execution loops, our agentic framework empowers the Controller to dynamically navigate the research space—backtracking or reflecting as needed. This autonomy allows the model to escape local optima that limit static workflows, resulting in ideas with significantly higher Novelty and Significance.

\textbf{Comparison with Backbone Model.} Comparison with the Qwen3-8B backbone highlights the critical role of active tool use. While the backbone relies solely on parametric knowledge, often leading to generic or unverified proposals, Agentic-Ideation leverages the Agentic Tool Space to ground reasoning in real-world literature. By verifying assumptions against retrieved evidence (e.g., via Get\_References), our agent significantly boosts Feasibility (e.g., +12.71\% in NLP), ensuring generated ideas are not just creative but technically sound.

\subsection{Efficiency of Data Synthesis}

Table ~\ref{table2} highlights the critical efficiency gap. Rejection Sampling proves prohibitively inefficient, averaging ~12 attempts to harvest a single valid sample. This stems from the vast, open-ended nature of scientific ideation, where unguided agents rarely converge spontaneously on the target idea, leading to massive computational waste.
In contrast, Agentic-Ideation achieves optimal efficiency, requiring only 1 rollout per sample. By leveraging our Oracle-Guided Strategy, the system uses "foresight" to ensure the necessary logic and tool invocations are generated in a single pass. Consequently, every generated trajectory is valid, delivering a >10$\times$ improvement in synthesis efficiency compared to rejection sampling, effectively making large-scale agentic data synthesis computationally feasible.

\begin{table}[H]
    \centering
    \caption{Comparison of the average number of rollout attempts required to synthesize a single valid training sample. "Roll out times" denotes the mean number of generation trials needed to produce one trajectory that passes the quality verification.}
    \label{table2}
    \begin{tabular*}{\linewidth}{l@{\extracolsep{\fill}}cccc}
        \toprule
        \textbf{Method} & \textbf{NLP} & \textbf{CV} & \textbf{Others} & \textbf{Overall} \\
        \midrule
        Reject Sampling & 14 & 9 & 13 & 12 \\
        Agentic-Ideation & 1 & 1 & 1 & 1 \\
        \bottomrule
    \end{tabular*}
\end{table}

\subsection{Ablation Study}

Table ~\ref{table3} investigates the individual contribution of each tool within our framework. The results unequivocally demonstrate that removing any single tool leads to a consistent degradation in performance across all metrics. This universal decline confirms the holistic nature of our Agentic Tool Space, where both external information acquisition (e.g., retrieval tools) and internal cognitive reasoning (e.g., reflection and gap analysis) are indispensable components. The synergy between these tools ensures that the agent can effectively bridge the gap between raw data and high-level scientific conceptualization.

\begin{table}[H]
    \vspace{-3pt}
    \caption{Ablation Study on the contribution of individual tools within the Agentic Tool Space.}
    \vspace{-6pt}
    \label{table3}
    \centering
        \begin{tabular*}{\linewidth}{l@{\extracolsep{\fill}}cccc}
            \toprule
            \textbf{Tools} & \textbf{Nov.} & \textbf{Sig.} & \textbf{Fea.} & \textbf{Overall} \\
            \midrule  
            w/o Search & 6.03 & 7.21 & 6.04 & 5.99\\
            w/o Get\_References & 6.34 & 7.29 & 5.98 & 6.32\\
            w/o Get\_Cited & 6.57 & 7.66 & 6.07 & 6.38\\
            w/o Analyse\_Gap & 5.86 & 7.26 & 5.99 & 6.07\\
            w/o Reflection & 6.49 & 7.34 & 6.10 & 6.43\\
            Full & 6.74 & 7.78 & 6.12 & 6.67\\
            \bottomrule
        \end{tabular*}
    \vspace{-5pt}
\end{table}

A closer inspection reveals that the Search and Analyse\_Gap tools are particularly critical. The exclusion of Search results in the lowest Overall score (5.99), indicating that without the ability to actively query external databases for broad information, the agent becomes confined to the limited context of initial references, severely hampering its ability to verify facts and expand its knowledge boundary. Similarly, removing Analyse\_Gap causes the most significant drop in the Novelty metric. This highlights the tool's pivotal role in the cognitive process; without explicitly synthesizing existing literature to pinpoint a "research void," the agent tends to generate ideas that are technically sound but incrementally trivial, failing to propose the distinct innovations that define high-quality research.

\subsection{Human Study}
To further validate the practical value and reliability of our method, we conducted a rigorous human evaluation. We adopted a double-blind protocol to eliminate potential bias. Expert evaluators were presented with anonymized ideas generated by different models in a randomized order and were asked to score them based on four criteria: Novelty, Significance, Feasibility, and Overall quality.

\begin{table}[H]
    \vspace{-3pt}
    \caption{Human evaluation results comparing Agentic-Ideation against baselines.}
    \vspace{-3pt}
    \label{table4}
    \centering
        \begin{tabular*}{\linewidth}{l@{\extracolsep{\fill}}cccc}
            \toprule
            \textbf{Methods} & \textbf{Nov.} & \textbf{Sig.} & \textbf{Fea.} & \textbf{Overall} \\
            \midrule 
            ResearchAgent & 5.83 & 6.41 & 5.41 & 5.57\\
            SciPIP & 6.02 & 6.49 & \underline{5.57} & \underline{5.87}\\
            Scimon & \underline{6.07} & \underline{6.68} & 5.55 & 5.76\\
            Virsci & 5.82 & 6.38 & 5.43 & 5.62\\
            Qwen3-8B & 5.96 & 6.57 & 5.42 & 5.55\\
            \midrule
            Agentic-Ideation & \textbf{6.71} & \textbf{7.50} & \textbf{6.05} & \textbf{6.45}\\
            \midrule
            \fontsize{10}{5}\selectfont Improvement\scalebox{1}{$\uparrow$} & 10.54\% &12.28\% &8.62\% &9.88\%\\ 
            \bottomrule
        \end{tabular*}
    \vspace{-5pt}
\end{table}

\begin{figure*}[h] 
\centering 

\newcommand{\stepsep}{\vspace{-5pt}\tcbline\vspace{-5pt}}

\begin{tcolorbox}[colback=gray!5, boxrule=0.8pt, title=\textbf{Generated Research Trajectory for "Adaptive Latent Remasking"}]
\fontsize{10}{11}\selectfont 

\textbf{Thought:} We've gathered initial papers, but I need to identify the common weaknesses across these approaches to pinpoint where the field is most vulnerable.\\
\tagAnalyse{Analyze\_Gap} Standard diffusion frameworks involve repeated stochastic denoising, where small errors amplify. A critical gap exists: models treat denoising as a fixed process without enabling dynamic correction of intermediate latent states...
\stepsep

\textbf{Thought:} I have identified the gap. Now I need to verify if recent works have bridged "self-guidance" with "latent refinement."\\
\tagGetRef{Get\_Cited} Query: ["Guiding a Diffusion Model with a Bad Version of Itself", ...]\\
\tagResult{Result} Found: ["Guiding a Diffusion Transformer with Internal Dynamics", ...]
\stepsep

\textbf{Thought:} "Internal Dynamics" suggests self-guidance is evolving. Now I need to check if discrete remasking (ReMDM) has been extended to continuous spaces.\\
\tagGetRef{Get\_Cited} Query: "Remasking Discrete Diffusion Models" \\
\tagResult{Result} Found: ["Corrective Diffusion Language Models"...]
\stepsep

\textbf{Thought:} "Corrective Diffusion" confirms post-training correction is possible in NLP. I should synthesize this with continuous latent spaces.\\
\tagSearch{Search} Query: "Corrective Diffusion Language Models Abstract"\\
\tagResult{Result} Abstract details: [proposes a correction-oriented post-training principle...]
\stepsep

\textbf{Thought:} Converging evidence suggests explicit error-correction is needed. I will propose "Latent-Space Remasking".\\
\tagIdeation{Ideation} (Draft 1) \textbf{Title:} Latent-Space Remasking. \textbf{Method:} Using a learned policy to remask low-confidence regions...
\stepsep

\textbf{Thought:} I've synthesized a unified mechanism. Now I must stress-test it to ensure it doesn't overlap with existing methods like RePaint.\\
\tagAnalyse{Reflection} The core mechanism of simple remasking closely mirrors existing work like "RePaint". \textbf{Decision:} REJECT. The idea lacks sufficient novelty. A refinement distinguishing it from fixed priors is required.
\stepsep

\textbf{Thought:} The Critic is right. The novelty should lie in \textit{adaptive} learning, utilizing internal error signals rather than fixed masks.\\
\tagIdeation{Ideation} (Final) \textbf{Title:} Adaptive Latent Refinement via Error-Aware Remasking. \textbf{Method:} We introduce a training method where the model learns a dynamic policy to identify and refine low-confidence latent regions based on internal error signals from the denoising trajectory, enabling self-driven correction without external conditioning.

\end{tcolorbox}
\caption{A visualized research trajectory generated by Agentic-Ideation. The agent actively explores the literature, identifies a gap in error correction, and crucially, utilizes the \textit{Reflection} tool to reject a redundant initial idea, leading to a more novel final proposal.}
\label{case_study}
\vspace{-6pt}
\end{figure*}

As shown in Table \ref{table4}, the human evaluation results align consistently with the automated metrics reported in the main experiment. Agentic-Ideation achieves superior performance across all dimensions, securing the highest Overall score of 6.45, which surpasses the strongest baseline (SciPIP) by 9.88\%. Notably, the most substantial improvement is observed in Significance (+12.28\%), suggesting that human experts particularly value the logical depth and theoretical value derived from our agent's grounded reasoning process. Furthermore, the +8.62\% gain in Feasibility confirms that the active usage of retrieval tools effectively reduces hallucinations, producing ideas that are not only innovative but also practically viable.

To intuitively demonstrate the working mechanism of Agentic-Ideation, we present a representative research trajectory in Figure \ref{case_study}. In this case, the agent explores the domain of "Diffusion Models" to address the problem of error compounding.

The trajectory illustrated in Figure \ref{case_study} offers a compelling validation of Agentic-Ideation's capabilities. Unlike standard LLMs that often generate generic or hallucinated limitations, our agent leverages the \tagAnalyse{Analyze\_Gap} tool to conduct a structural critique of diffusion models, identifying the specific theoretical void of inference-time correction rather than superficial issues. Throughout the process, the agent demonstrates strategic information seeking by actively linking distinct sub-fields—tracing "Self-Guidance" via \tagGetRef{Get\_Cited} and deep-diving into "Corrective Diffusion" via \tagSearch{Search}—to synthesize a hybrid hypothesis that bridges discrete text correction with continuous image generation. Most crucially, the \tagAnalyse{Reflection} step acts as a rigorous quality filter; when the agent detects that its initial "Latent-Space Remasking" proposal lacks sufficient novelty compared to prior work like RePaint, it autonomously rejects the draft and iterates. This self-correction loop ensures that the final output, "Adaptive Error-Aware Learning," is not only scientifically grounded but also distinctly innovative, demonstrating that our Agentic Tool Space serves as both a generative engine and an effective quality gatekeeper.

%% file: 5.Conclusion.tex
\section{Conclusion}
In this work, we address the rigidity of workflow-based approaches and the inefficiency of agentic data synthesis for scientific ideation. We introduce Agentic-Ideation, an Oracle-Guided framework that efficiently navigates the vast search space, bypassing the computational bottlenecks of naive exploration. By training on these synthesized trajectories, our model acquires robust capabilities for grounded reasoning and autonomous exploration. Empirical results confirm that Agentic-Ideation establishes state-of-the-art performance while boosting synthesis efficiency by over 10$\times$.

%% file: 6.appendix.tex
\newpage
\appendix
\section{Appendix}
\subsection{More Details of Evaluation Metrics}
\label{metrics}

To ensure a rigorous and consistent assessment of the generated research ideas, we employ a standardized scoring rubric ranging from 1 to 10 across four key dimensions: Novelty, Significance, Feasibility, and Overall Score. Both the automated LLM judges and human evaluators adhere to the detailed criteria outlined below.

\textbf{Novelty (1-10)}
Evaluates the innovation of the proposed method.
\begin{itemize}[left=0pt] 
    \item \textbf{9-10 (Groundbreaking):} Proposes a completely new paradigm, mathematical formulation, or solves a long-standing open problem in a unique way.
    \item \textbf{7-8 (Significant):} Introduces a clever variation or a non-obvious combination of existing techniques that offers substantial new insights.
    \item \textbf{5-6 (Incremental):} Represents a standard application of known methods to a new domain or offers only minor improvements.
    \item \textbf{1-4 (Trivial):} Lacks originality; constitutes derivative work or merely repeats a known solution.
\end{itemize}

\textbf{Significance (1-10)}
Evaluates the potential impact and importance of the addressed problem.
\begin{itemize}[left=0pt] 
    \item \textbf{9-10 (High Impact):} Addresses a critical bottleneck in the field; results would likely influence future research directions.
    \item \textbf{7-8 (Important):} Solves a meaningful problem within a specific subfield; highly relevant to the community.
    \item \textbf{5-6 (Moderate):} The problem is valid but niche or of limited interest to the broader community.
    \item \textbf{1-4 (Low):} The problem is unimportant, contrived, or already well-solved.
\end{itemize}

\textbf{Feasibility (1-10)}
Evaluates the technical soundness and likelihood of success.
\begin{itemize}[left=0pt] 
    \item \textbf{9-10 (Very High):} The methodology is rigorous, assumptions are clearly valid, and the proposed steps are logical and fully implementable.
    \item \textbf{7-8 (High):} The approach is sound, though some minor implementation details might need further clarification.
    \item \textbf{5-6 (Uncertain):} There are theoretical gaps or risky assumptions that might lead to failure.
    \item \textbf{1-4 (Low):} Fundamentally flawed; the method contradicts established principles or is impossible to implement.
\end{itemize}

\textbf{Overall Score (1-10)}
A holistic assessment of the idea's quality.
\begin{itemize}[left=0pt] 
    \item \textbf{9-10 (Excellent):} An outstanding idea with clear potential to become a seminal work. Extremely promising.
    \item \textbf{7-8 (Good):} A strong, solid idea that is well-conceived and worth pursuing.
    \item \textbf{5-6 (Fair):} Has some merit but contains significant flaws, gaps, or lack of depth that need to be addressed.
    \item \textbf{1-4 (Poor):} Critically flawed, lacks substance, or is not ready for serious consideration.
\end{itemize}

\subsection{Compute Sources}
\label{implementation}

All experiments were conducted on a high-performance computing cluster. Table \ref{table:hyperparameters} summarizes the detailed computing resources, data statistics, and training hyperparameters used for fine-tuning the Agentic LLM.

\begin{table}[h]
    \centering
    \caption{\textbf{Implementation Details and Hyperparameters.} The model was fully fine-tuned on the synthesized trajectory dataset.}
    \label{table:hyperparameters}
    \begin{tabular*}{\linewidth}{l@{\extracolsep{\fill}}r}
        \toprule
        \textbf{Configuration} & \textbf{Value} \\
        \midrule
        \multicolumn{2}{l}{\textit{Compute \& Data}} \\
        \midrule
        Computing Infrastructure & 4 $\times$ NVIDIA A100 (80GB) \\
        Backbone Model & Qwen3-8B \\
        Tuning Strategy & Full Fine-Tuning \\
        Training Data Size & 4,646 Trajectories \\
        Total Training Time & $\sim$ 10.5 Hours \\
        \midrule
        \multicolumn{2}{l}{\textit{Training Parameters}} \\
        \midrule
        Number of Epochs & 5 \\
        Initial Learning Rate & 5e-5 \\
        LR Scheduler Type & Cosine \\
        \bottomrule
    \end{tabular*}
\end{table}

\subsection{Limitations}
Despite the promising results, we acknowledge two primary limitations. First, our experiments rely on the Qwen3-8B backbone; while efficient, its reasoning depth and world knowledge are naturally constrained compared to larger-scale foundation models, suggesting that scaling up could further enhance idea complexity. Second, the current agentic tool space is predominantly retrieval-oriented, focusing on information acquisition rather than executable actions. Future work should integrate "active" tools, such as Python code execution or domain-specific simulators, to allow the agent to perform preliminary validation of its proposed ideas.